\documentclass[fleqn,10pt]{SelfArx} % Document font size and equations flushed left

\usepackage[english]{babel} % Specify a different language here - english by default

\usepackage{textcomp}
\usepackage[font=small]{caption}

\definecolor{color1}{RGB}{0,0,90} % Color of the article title and sections
\definecolor{color2}{RGB}{0,20,20} % Color of the boxes behind the abstract and headings

\usepackage{hyperref} % Required for hyperlinks
\hypersetup{hidelinks,colorlinks,breaklinks=true,urlcolor=color2,citecolor=color1,linkcolor=color1,bookmarksopen=false,pdftitle={Title},pdfauthor={Author}}

\JournalInfo{~} % Journal information
\Archive{~} % Additional notes (e.g. copyright, DOI, review/research article)

\PaperTitle{Deep-learning-based identification of odontogenic keratocysts in hematoxylin- and eosin-stained jaw cyst specimens} % Article title

\Authors{Kei Sakamoto\textsuperscript{1}*, Kei-ichi Morita\textsuperscript{2}, Tohru Ikeda\textsuperscript{1}, Kou Kayamori\textsuperscript{1}} % Authors
\affiliation{\textsuperscript{1}\textit{Department of Oral Pathology, Tokyo Medical and Dental University, Tokyo, Japan}} % Author affiliation
\affiliation{\textsuperscript{2}\textit{Department of Maxillofacial Surgery, Tokyo Medical and Dental University, Tokyo, Japan}} % Author affiliation
\affiliation{*\textbf{Corresponding author}: s-kei.mpa@tmd.ac.jp} % Corresponding author

 % Keywords - if you don't want any simply remove all the text between the curly brackets
\newcommand{\keywordname}{Keywords} % Defines the keywords heading name

\Abstract{The aim of this study was to develop a digital histopathology system for identifying odontogenic keratocysts in hematoxylin- and eosin-stained tissue specimens of jaw cysts. Approximately 5000 microscopy images with 400$\times$ magnification were obtained from 199 odontogenic keratocysts, 208 dentigerous cysts, and 55 radicular cysts. A proportion of these images were used to make training patches, which were annotated as belonging to one of the following three classes: keratocysts, non-keratocysts, and stroma. The patches for the cysts contained the complete lining epithelium, with the cyst cavity being present on the upper side. The convolutional neural network (CNN) VGG16 was finetuned to this dataset. The trained CNN could recognize the basal cell palisading pattern, which is the definitive criterion for diagnosing keratocysts. Some of the remaining images were scanned and analyzed by the trained CNN, whose output was then used to train another CNN for binary classification (keratocyst or not). The area under the receiver operating characteristics curve for the entire algorithm was 0.997 for the test dataset. Thus, the proposed patch classification strategy is usable for automated keratocyst diagnosis. However, further optimization must be performed to make it suitable for practical use.}

\begin{document}

\flushbottom % Makes all text pages the same height

\maketitle % Print the title and abstract box

%\tableofcontents % Print the contents section

\pagestyle{empty} % Removes page numbering from the first page
\rhead{}
\pagestyle{fancy}
%\lhead{Identification of odontogenic keratocysts using deep learning}
\cfoot{\thepage}

%----------------------------------------------------------------------------------------
%	ARTICLE CONTENTS
%----------------------------------------------------------------------------------------

\section*{Introduction} % The \section*{} command stops section numbering
Many attempts have been made over the years to use automated machine vision systems for the analysis of medical images~\cite{bib1, bib2}. Although the development of machine vision systems in the field of histopathology has generally lagged, research efforts to develop practical systems in this field are now accelerating~\cite{bib3, bib4}. An automated system is expected to reduce the workload of pathologists and make up for the shortage in experts in the field~\cite{bib1}. Pathologists analyze abnormal histological findings in tissue specimens taken from patients and interpret their findings to deduce the etiology. These findings are then integrated to provide conclusive remarks in the form of a pathological diagnosis. The process calls for both expert knowledge and practical experience. However, if one only focuses on the source and the end product, making a pathological diagnosis can be regarded as an image classification task that incorporates additional information such as the site of the tissue, the patient's clinical history, and the macroscopic appearance of the tissue.

In machine learning, classification is done based on the optimized boundary for discrimination in a high-dimensional feature space~\cite{bib5}. For this purpose, discriminant features are extracted from images and assessed statistically. The features used conventionally in the automated analysis of histopathological images include generalized textural ones such as the contrast or entropy, as well as more specialized cytological features such as the number, size, and distribution of the cell nuclei or cell borders~\cite{bib1}. These approaches are suitable for solving a few specific problems. However, there exists a gap between the research being done and its applicability in real-world pathological investigations. Until the early 2010s, extracting useful features was a laborious task and had to be peformed manually by human researchers. However, deep learning using convolutional neural networks (CNNs) has rapidly become the method of choice in image analysis. In this case, computers are trained using example data, and one can autonomously optimize the parameters. With respect to the machine learning analysis of histologic images, there exist three major challenges: (1) the detection and evaluation of cellular structures such as nuclei, (2) the detection, evaluation, and segmentation of tissue structures such as glands and tumor nests, and (3) the classification of the entire slide image. There have been several studies on analyzing images related to major diseases such as breast cancer~\cite{bib6, bib7, bib8}, prostate cancer~\cite{bib9, bib10}, and colon cancer~\cite{bib11, bib12}, which affect a large number of people around the world. In contrast, rare diseases have seldom been the focus of this approach, mainly owing to the difficulty in collecting large sets of related images. Still, the diagnosis of rare diseases is a promising area of application for machine vision systems, as such systems should be able to assist pathologists in evaluating unfamiliar diseases.

Odontogenic keratocysts (referred to as keratocysts hereafter) are relatively rare cysts that form in the jaw bone, accounting for approximately 10\% of all jaw cysts~\cite{bib13}. Keratocysts are a distinct entity with a unique histologic appearance. The lining epithelium is a squamous epithelium composed of approximately 5–8 layers of cells, whose surface is parakeratinized. The basal cells are cuboidal or columnar with elliptical nuclei and are aligned regularly, generating a so-called palisading pattern~\cite{bib14}. In other types of jaw cysts, the lining epithelium does not exhibit the distinct histologic characteristics seen in the case of keratocysts. Thus, jaw cysts other than keratocysts (referred to as non-keratocysts hereafter) are defined by their location and their association with a dental treatment or inflammation instead of based on histological characteristics. These include radicular cysts and dentigerous cysts, which account for approximately 50\% and 20\%, respectively, of all jaw cysts~\cite{bib13, bib14}. Non-keratocysts are treated by enucleation. However, different treatment strategies are used in the case of keratocysts, because they show more aggressive growth, exhibiting high recurrence rates if treated simply by enucleation~\cite{bib15, bib16}. Therefore, the precise classification of keratocysts and non-keratocysts is crucial with respect to the pathological investigations of jaw cysts. Since there are no definitive histologic features for non-keratocysts, the challenge is to be able to distinguish keratocysts from the other types of jaw cysts.

Although jaw cysts are observed relatively frequently in medical departments that deal with oral lesions, they are less frequently encountered by general pathologists. Furthermore, the histologic appearance of keratocysts varies across cases, and there are no quantitative criteria to evaluate the related features. This variety and ambiguity results in considerable subjectivity in decision making, often leading to confusion and difficulties with respect to reproducibility, particularly in the case of inexperienced observers.

Several researchers have performed digital image analysis using microscopy images of keratocysts with 400$\times$ magnification. It has been reported there exist quantitative differences in the morphometric parameters of keratocysts and radicular cysts~\cite{bib17, bib18, bib19}. Since the algorithms employed in these studies were optimized and tested using a relatively small number of samples (approximately 20 cases of keratocysts), it remains unclear whether these algorithms can be used for real-world pathological analysis. Further, the algorithms used were conventional feature engineering algorithms, and a deep-learning-based approach that uses a larger dataset may improve the precision of the classification process and hence increase the suitability of this approach for real-world applications.

To reduce the subjectivity of the classification process and aid pathologists, we attempted to develop a histopathologic image analysis system for keratocyst detection. Since keratocyst specimens are submitted to the laboratory with prior labeling as to whether they were taken from a jaw cyst, the aim of this work was set to develop an algorithm that can determine whether a microscopy image of a jaw cyst specimen is that of a keratocyst or not. The ultimate goal is to be able to analyze and classify the entire slide image. However, in this study, we only analyzed high-magnification (400$\times$) microscopy images.

\addcontentsline{toc}{section}{Introduction} % Adds this section to the table of contents

%------------------------------------------------

\section{Methods}
\subsection*{Specimens}
Tissue specimens corresponding to 199 cases of odontogenic keratocysts, 208 cases of dentigerous cysts, and 55 cases of radicular cysts were collected from the archives of the Dental Hospital of Tokyo Medical and Dental University. The tissues were fixed in 10\% buffered formalin and embedded in paraffin as per the routine laboratory protocol. Next, 4-mm-thick sections were cut and stained with hematoxylin and eosin (H\&E). This work was approved by the Ethics Committee of Tokyo Medical and Dental University (Registry Number D2018-061).

\subsection*{Computational framework}
All the computational analyses were performed on a GPU (GeForce 1080Ti, Nvidia, Santa Clara, CA, USA)-equipped personal computer (Intel\textregistered Core\texttrademark i7-8700 3.2 GHz, 64 GB RAM, Windows 10). The neural networks were built using Keras~\cite{bib20}.

\subsection*{Image datasets}
The histological images were obtained using an optical microscope (BX43, Olympus, Tokyo, Japan) at 400$\times$ magnification, a digital camera (DP70, Olympus), and the imaging software CellSense (Olympus). The automatic exposure mode was used for exposure compensation. The maximum number of images taken from one specimen was limited to 30 in order to prevent overrepresentation. In the case of keratocysts (referred to as \textbf{K} hereafter), the areas that contained the lining epithelium with characteristic histologic features were located, and 2412 photographs were taken such that the images did not overlap. A total of 2293 photographs of dentigerous cysts were collected as the primary type of non-keratocysts (referred to as \textbf{N} hereafter). For the images of the stroma (referred to as \textbf{S} hereafter), 210 photographs were collected from 30 jaw cyst specimens at the areas that lacked the lining epithelium and were composed solely of stromal tissues that were fibrous or inflammatory granulation tissues. A total of 50 photographs of radicular cysts were collected for the test dataset (referred to as \textbf{R} hereafter)(Figure 1A).

The photographs were saved as 1224$\times$960 pixel (px) image files in the jpeg format. These images are referred to as ``\textit{large images}" hereafter. During preprocessing, all \textit{large images} were normalized using the Match Color function of Photoshop CS4 (Adobe, San Jose, CA, USA).

Next, 1704 \textit{large images} of \textbf{K} corresponding to 102 cases and 1635 \textit{large images} of \textbf{N} corresponding to 105 cases were used to make the training patches (256$\times$256 px) ; this procedure is discussed later (Figure 1B).

These image patches are referred to as ``\textit{small images}" hereafter. Next, 160 \textit{large images} of 19 jaw-cyst-derived specimens that did not contain the lining epithelium were used to generate 5000 \textit{small images} of \textbf{S}. An area of 362$\times$362 px was randomly cropped from \textit{large images} and rotated at a random angle. Then, the central 256$\times$256 px area was removed. 
In addition, 658 \textit{large images} of \textbf{K} from 72 cases and 608 \textit{large images} of \textbf{N} from 78 cases were used for training the second CNN (Figure 1A). In addition, 100 \textit{small images} each of \textbf{K} and \textbf{N} were collected from this dataset using the first method (Figure 1B) described in the Results section to test the first CNN. A total of 50 \textit{large images} of \textbf{K} from 25 cases, 50 \textit{large images} of \textbf{N} from 25 cases, 50 \textit{large images} from 11 jaw-cyst-exhibiting specimens that did not contain the lining epithelium (i.e., \textbf{S}), and 50 \textit{large images} of \textbf{R} from 25 cases were used to test the entire model.
\textit{Small images} of \textbf{K} and \textbf{N} were obtained by manually cropping \textit{large images}. The patch locations were such that they contained the definitive histologic features necessary and sufficient for classification. 

In specimens in which the lining epithelium was too thick to fit within an area of 256$\times$256 px, the patch size was increased to 320$\times$320 px, and the cropped images were resized to 256$\times$256 px. These downsized images were less than 2\% of the total number of \textit{small images}. Samples with thicker lining epithelia whose layers were larger than the 320$\times$320 px patch were omitted. Several \textit{small images} were from individual \textit{large images}; however, care was taken to avoid overlapping.

\subsection*{Training of models}
VGG16~\cite{bib21} or Resnet34~\cite{bib22} was implemented for the first CNN. For VGG16, parameters obtained using ImageNet~\cite{bib23} were used as the initial values, and the parameters for the bottom 10 layers were determined during training. \textit{Small images} were fed into the first CNN after they had been converted into 224$\times$224 px images to fit the Resnet34 architecture as well as to reduce the computation load. The hyperparameters were determined empirically. \textit{Large images} were resized to 1071$\times$840 px before being scanned and analyzed by the first CNN. The 224$\times$224 px patch of a 1071$\times$840 px \textit{large image} corresponded to the 256$\times$256 px patch on a 1224$\times$960 px \textit{large image}. A simple CNN with three convolutional layers was employed as the second CNN. Training was performed thrice, and the highest accuracy and lowest loss in validation data were recorded during each training step, and the mean values were considered.

\begin{figure}[!hbt]
\captionsetup{format = plain, justification=justified}
\includegraphics[width=9cm, bb=0 -50 1440 1728]{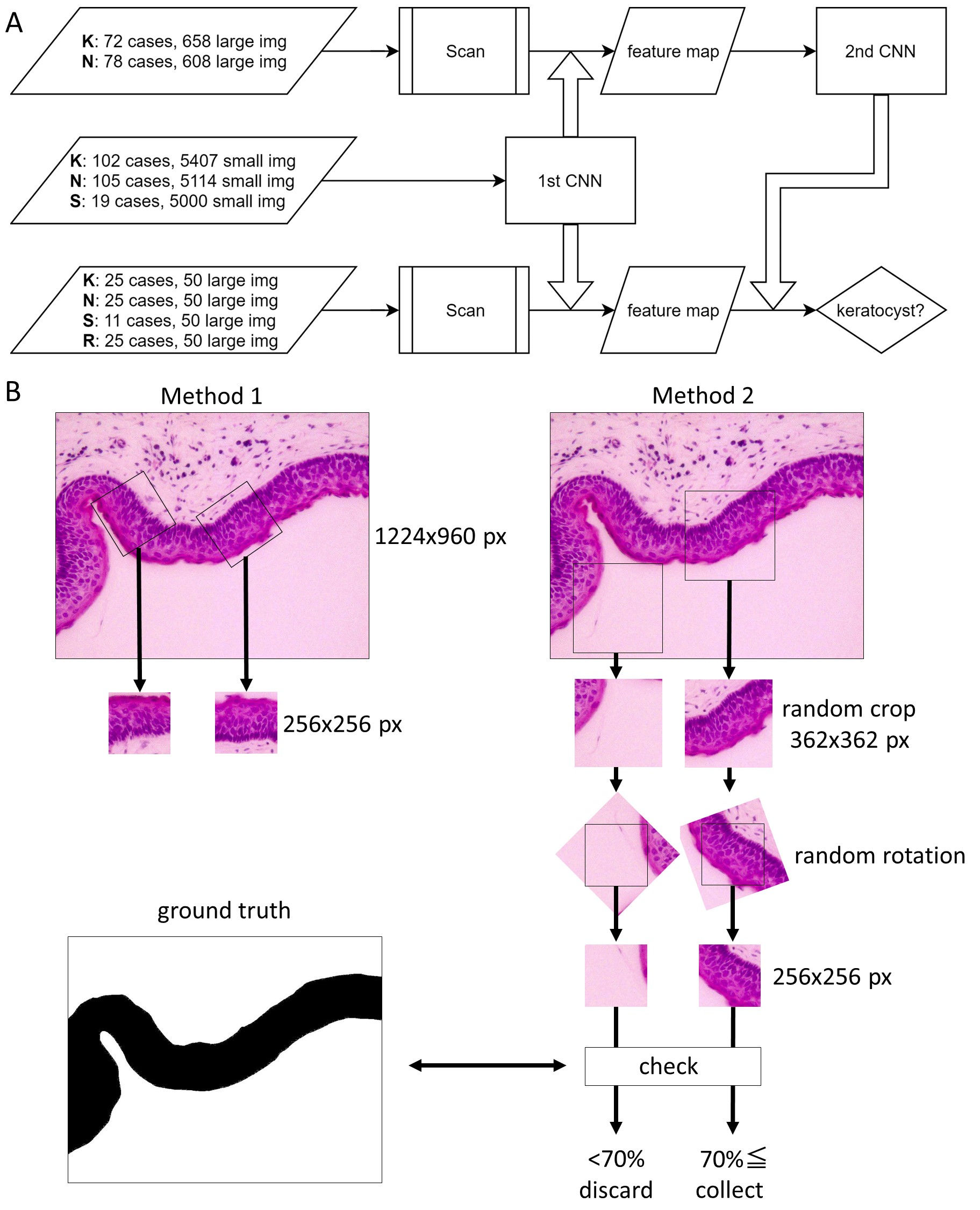}
\caption{Workflow employed. A) Three independent sets of cases were used to train first and second CNNs and to test entire model. ``Large img" denotes 1224$\times$960 px images obtained at magnification of 400$\times$. ``Small img" denotes 256$\times$256 px image patches. Arrows represent input or processing. Big arrows represent analysis. \textbf{K}: keratocysts, \textbf{N}: non-keratocysts, \textbf{S}: stromal tissues, and \textbf{R}: radicular cysts. B) Methods for generating patches. In first method, 256$\times$256 px patches were manually located on lining epithelium and cropped. In all patches, cyst cavity is located on upper side. In second method, ground truth images were prepared manually. 362$\times$362 px patches were randomly cropped and rotated at random angle. Then, central 256$\times$256 px portion was cropped. Corresponding ground truth images was referred to and patches in which keratocyst elements made up less than 70\% of image were discarded.}
\label{fig1:view}
\end{figure}

%------------------------------------------------

\section{Results}
First, we prepared a training image dataset of keratocysts (\textbf{K}) and non-keratocysts (\textbf{N}). We assumed that the most difficult step in this task would be discriminating between \textbf{K} and the dentigerous cysts with thick lining epithelia. The latter are sometimes similar to radicular cysts in appearance, once inflammation has subsided, and histologically hard to distinguish from dentigerous cysts (Figure 2). Therefore, we collected images of \textbf{N} that included those of dentigerous cysts. From \textit{large images} (1224$\times$960 px), patches of \textit{small images} (256$\times$256 px) were cropped using two different methods.

\begin{figure}[!hbt]\centering
\captionsetup{format = plain, justification=justified}
\includegraphics[width=8cm, bb=0 -15 914 737]{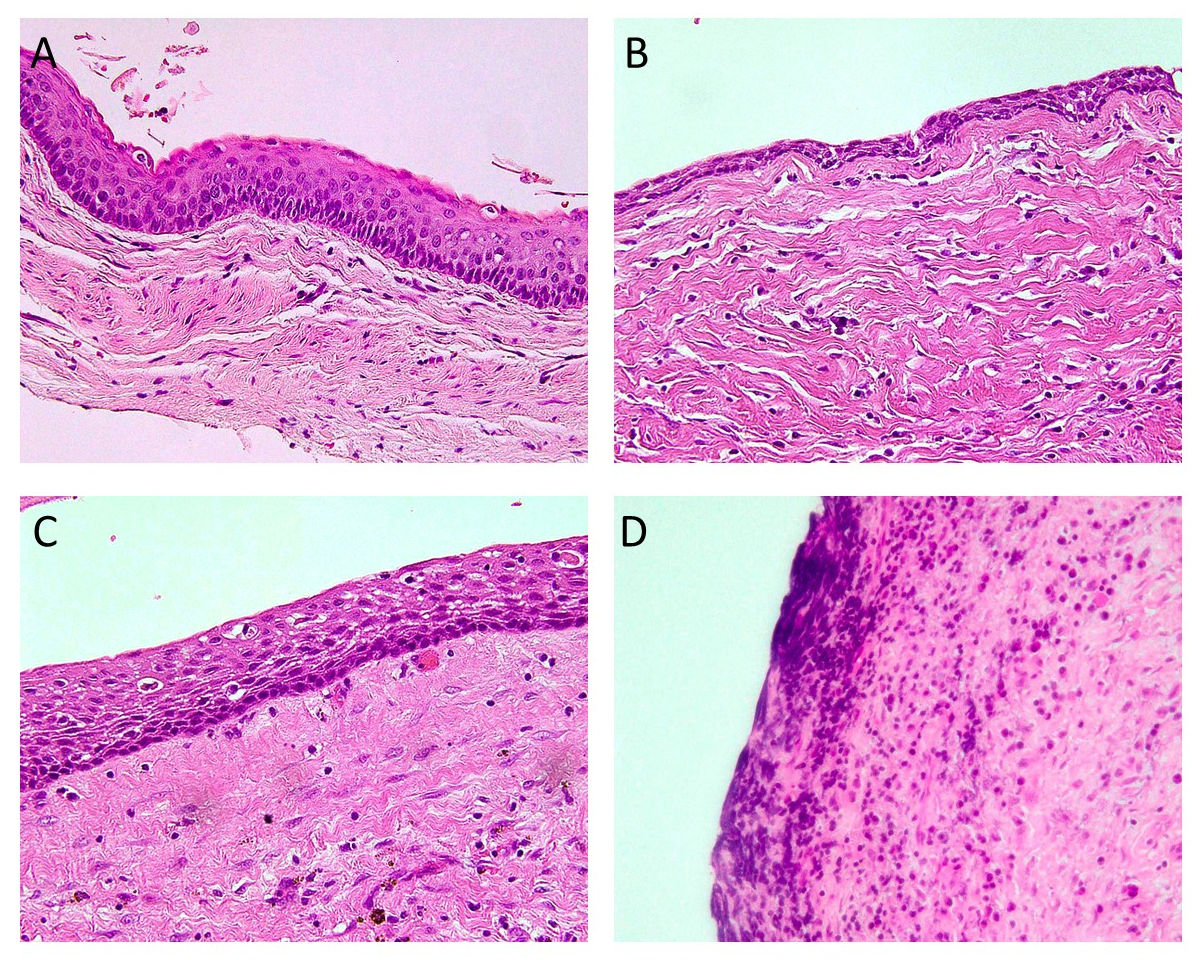}
\caption{Representative histology of jaw cysts. A) Keratocyst. B) Dentigerous cyst. C) Dentigerous cyst with thick epithelium. D) Local image of radicular cyst with lining epithelium missing.}
\label{fig2:view}
\end{figure}

In the first method, \textit{small images} were manually cropped and collected. The patch size was determined empirically such that the entire lining epithelium fit within a patch of this size in the case of most samples. The lining epithelium has a cavity side and a stromal side. To preserve this orientation in the image data, the training images were cropped after adjusting the rotation so that the cyst cavity appeared on the upper side of the patch (Figure 1B). Dataset \textbf{K} was created to be representative of the typical histology (Figure 2A) in the same proportion as is the case for the entire population. For \textbf{N}, the common histologic appearance was a thin lining epithelium consisting of 2–3 cell layers (Figure 2B). The majority of the samples in \textbf{N} with this histological feature could be distinguished from those in \textbf{K} with ease; the bigger challenge was to distinguish the samples in \textbf{N} with a thick lining epithelium (Figure 2C). For efficient training, the latter samples in \textbf{N} were collected in case they contained the thick epithelium. Images showing a common \textbf{N}-like histology were also included in the dataset but in a proportion smaller than that for the entire population. Therefore, the collected \textit{small images} of \textbf{N} were not a subset of a statistical sample reflecting the entire population but were biased towards the cases that were difficult to discriminate from those in \textbf{K}. The training dataset consisted of 5407 and 5114 \textit{small images} corresponding to \textbf{K} and \textbf{N}, respectively (Figure 1A). In the second method, we manually annotated the main body of the samples in \textbf{K}, that is, the lining epithelium, and formed ground truth images (Figure 1B). Then, 362$\times$362 px patches were obtained by cropping \textit{large images} at random locations. The cropped images were then rotated by a random angle, and the central 256$\times$256 px portion was cropped. Patches wherein more than half of the total area was blank (i.e., no tissue present) were discarded. For the images of \textbf{K}, the teacher data images were referred to, and only the images where the \textbf{K}-related elements made up more than 70\% of the total area were considered (Figure 1B). A total of 10000 \textit{small images} each of \textbf{K} and \textbf{N} were collected using the second method.

We adopted the VGG16 and Resnet34 CNN architectures. Both VGG16 and Resnet34 were fine tuned. The \textit{small image} dataset was fed into both networks after the images had being resized to 224$\times$224 px, so that they matched the Resnet architecture.

The \textit{small image} dataset was divided into training and validation datasets in a 20:1 ratio, and both CNNs were trained to achieve an accuracy of more than 98\% for the validation dataset. The trained CNNs were evaluated using a \textit{small image} test dataset that consisted of 100 \textbf{K} and 100 \textbf{N} samples.  The area under the receiver operating characteristic (ROC) curve (AUC) value was 0.998 for VGG16 and 0.996 for Resnet34 (Table 1), when the CNNs were trained using the dataset prepared manually using the first method (Figure 1B).

\begin{table}[!hbt]
\centering
\captionsetup{format = plain, justification=justified}
\begin{tabular}{lcccc}
\toprule
%\multicolumn{2}{c}{Name} \\
%\cmidrule(r){1-2}
{} & \textbf{K}/\textbf{N} & \textbf{K}/\textbf{N}/\textbf{S} & \textbf{K}/\textbf{(N+S)} & \textbf{K}/\textbf{N} method 2 \\
\midrule
VGG16 & $0.998$ & $0.998$ & $0.998$ & $0.860$ \\
VGG16 init & $0.997$ & $0.978$ & $0.995$ & $0.836$ \\
Resnet34 & $0.996$ & $0.990$ & $0.998$ & $0.865$ \\

\bottomrule\\
\end{tabular}
\caption{AUC values of first CNN tested using dataset consisting of 100 keratocysts and 100 non-keratocysts. For ``VGG16 init", kernel initialization was performed using trainable filters before retuning.  \textbf{K}/\textbf{N} denotes binary classification between keratocysts and non-keratocysts. \textbf{K}/\textbf{N}/\textbf{S} denotes three-category classification and includes the stroma. \textbf{K}/\textbf{(N+S)} denotes binary classification between keratocysts and non-keratocysts, including the stroma. \textbf{K}/\textbf{N} method 2 means model was trained using patch dataset prepared using second method (see Figure 1B).}
\label{tab:label}
\end{table}

Thus, both models were found to be promising classifiers for \textbf{K} and \textbf{N} (Figures 3A and 3B). The models trained using the dataset prepared by the second method (Figure 1B) exhibited much lower AUC values (Table 1), which did not improve significantly after hyperparameter tuning. Hence, we decided to use the first CNN trained using the manually prepared dataset for the rest of the study.

\begin{figure}[!hbt]\centering
\captionsetup{justification=justified}
\includegraphics[width=6cm, bb=115 -20 872 860]{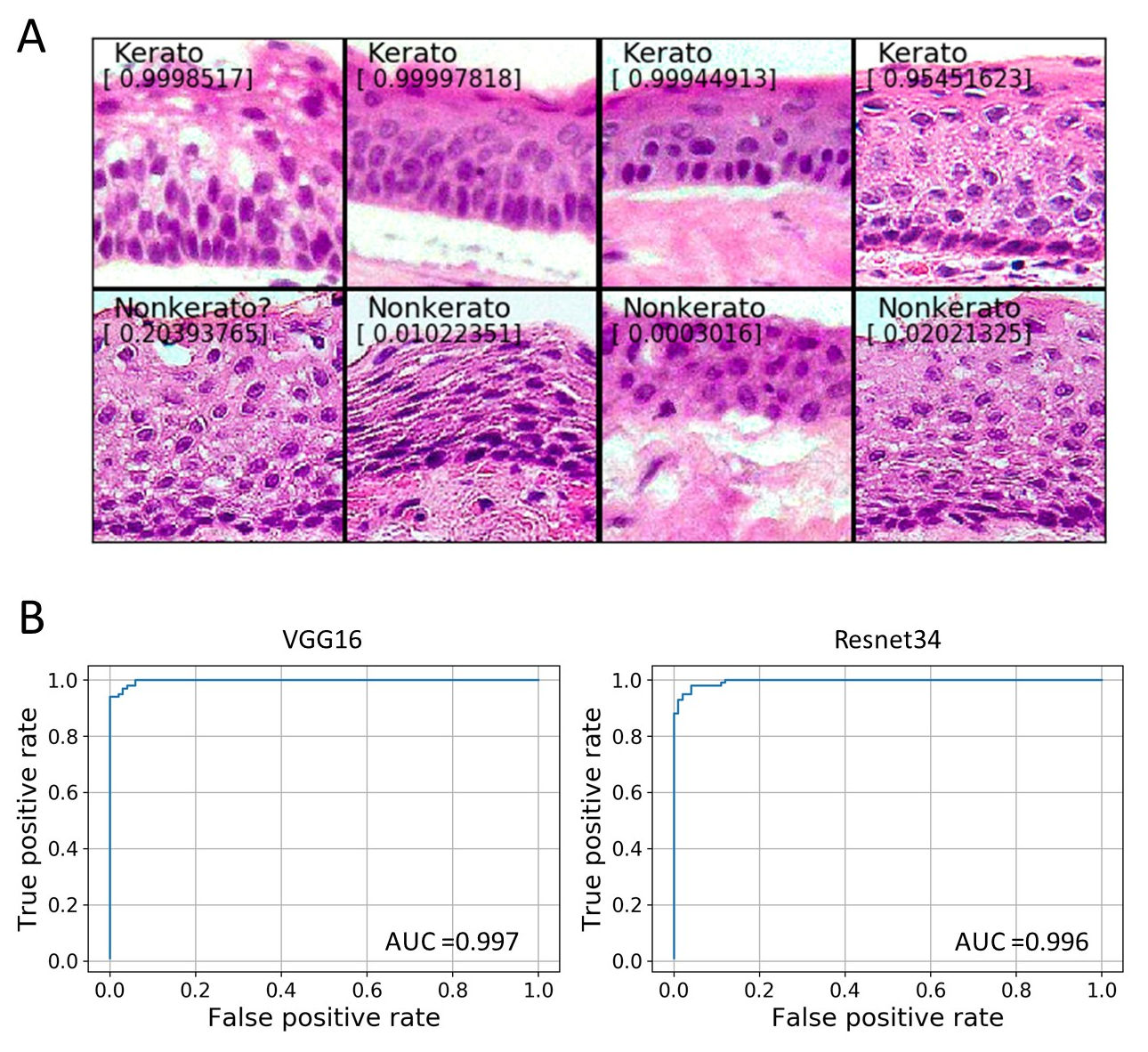}
\caption{Results for first CNN (VGG16) trained using dataset prepared by manual cropping. A) Selected \textit{small images} of keratocysts and non-keratocysts superimposed with probabilities of keratocysts and predictions (threshold = 0.5). First column shows keratocysts and second column shows non-keratocysts. B) ROC curves and AUCs of first CNNs (VGG16 and Resnet34) as determined using dataset containing 100 images each of keratocysts and non-keratocysts.}
\label{fig3:view}
\end{figure}

To gain insights into the image features based on which the model discriminates between \textbf{K} and \textbf{N}, we checked the outputs of the VGG16 convolutional filters. Of the 512 filters that constituted the last convolutional layer, 60–70\% showed outputs that were zero in response to \textbf{K} and \textbf{N} (Figure 4A). Although the list of the activated filters is shared by \textbf{K} and \textbf{N}, some filters were significantly more activated by \textbf{K} (Figure 4B), implying that these filters are important for recognizing \textbf{K}. To further examine the filter functions, input images that maximized the output of the convolutional filters were generated using the gradient ascent method~\cite{bib24}. The filters that were activated by the \textbf{K} and \textbf{N} images generated only a few types of patterns using the gradient ascent method. The most common pattern was an ``aligned capsule"-like pattern (Figure 4C). Filter activation maps superimposed onto the input image~\cite{bib25} revealed that the locations for filter activation were the areas that included the basal layer of \textbf{K} (Figure 4D). Another type of filter responded to spinous cells with prominent nucleoli (Figure 4C).

\begin{figure}[!hbt]\centering
\captionsetup{format = plain, justification=justified}
\includegraphics[width=7cm, bb=0 -20 1138 1484]{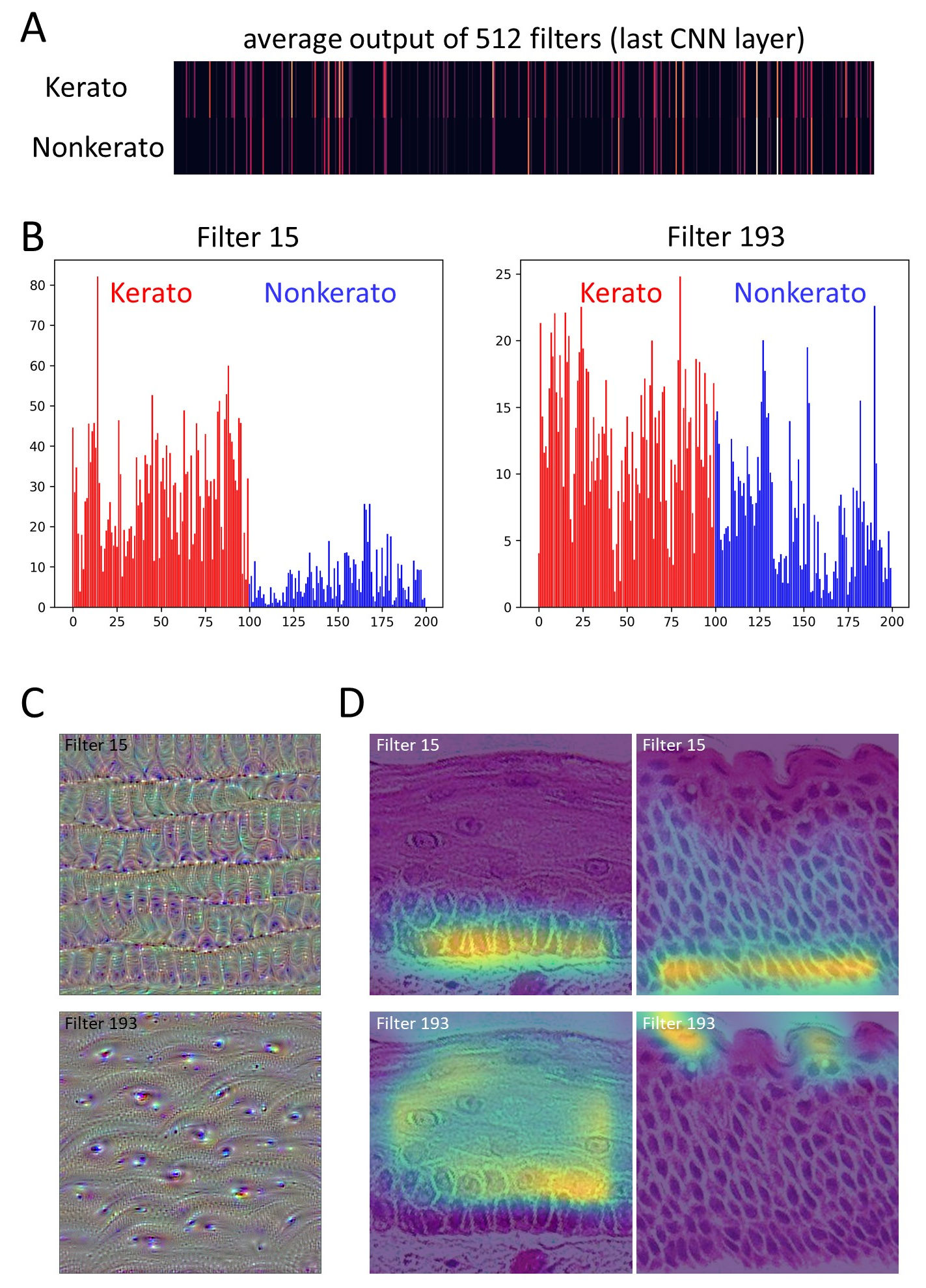}
\caption{Visualization of last convolutional layer of VGG16. A) Landscape of outputs from last convolutional layer. Outputs were averaged on spatial plane, and means of three outputs in response to three different images are shown as heatmap. There are 512 filters, aligned horizontally, in the heatmap. Activated filters are represented by lines whose brightness corresponds to output value, whereas black areas are indicative of zero or nearly zero outputs from the filters. B) Bar charts of outputs from representative filters in response to test dataset. Horizontal axis represents sample number and vertical axis represents output value.  Numbers 0 to 99 are keratocysts. Numbers 100 to 199 are non-keratocysts. Left; filter 15. Right; filter 193. C) Representative filter patterns generated by gradient ascent method. D) Filter activation heatmaps superimposed onto input images of keratocysts.}
\label{fig4:view}
\end{figure}

The facts that most filters were not used and that similar ones were redundant may be attributed to the suboptimal kernel initialization of the trainable layers because the values obtained through training using ImageNet~\cite{bib23} data were used as the initial values~\cite{bib20}. When these filters were initialized using a uniform or Gaussian distribution, most of them returned interpretable images with various patterns when the gradient ascent method was employed (Figure 5). However, the accuracy of classification was lower as compared to that for VGG16 without kernel initialization (Table 1).

\begin{figure}[!hbt]\centering
\captionsetup{format = plain, justification=justified}
\includegraphics[width=9cm, bb=0 -20 1365 1200]{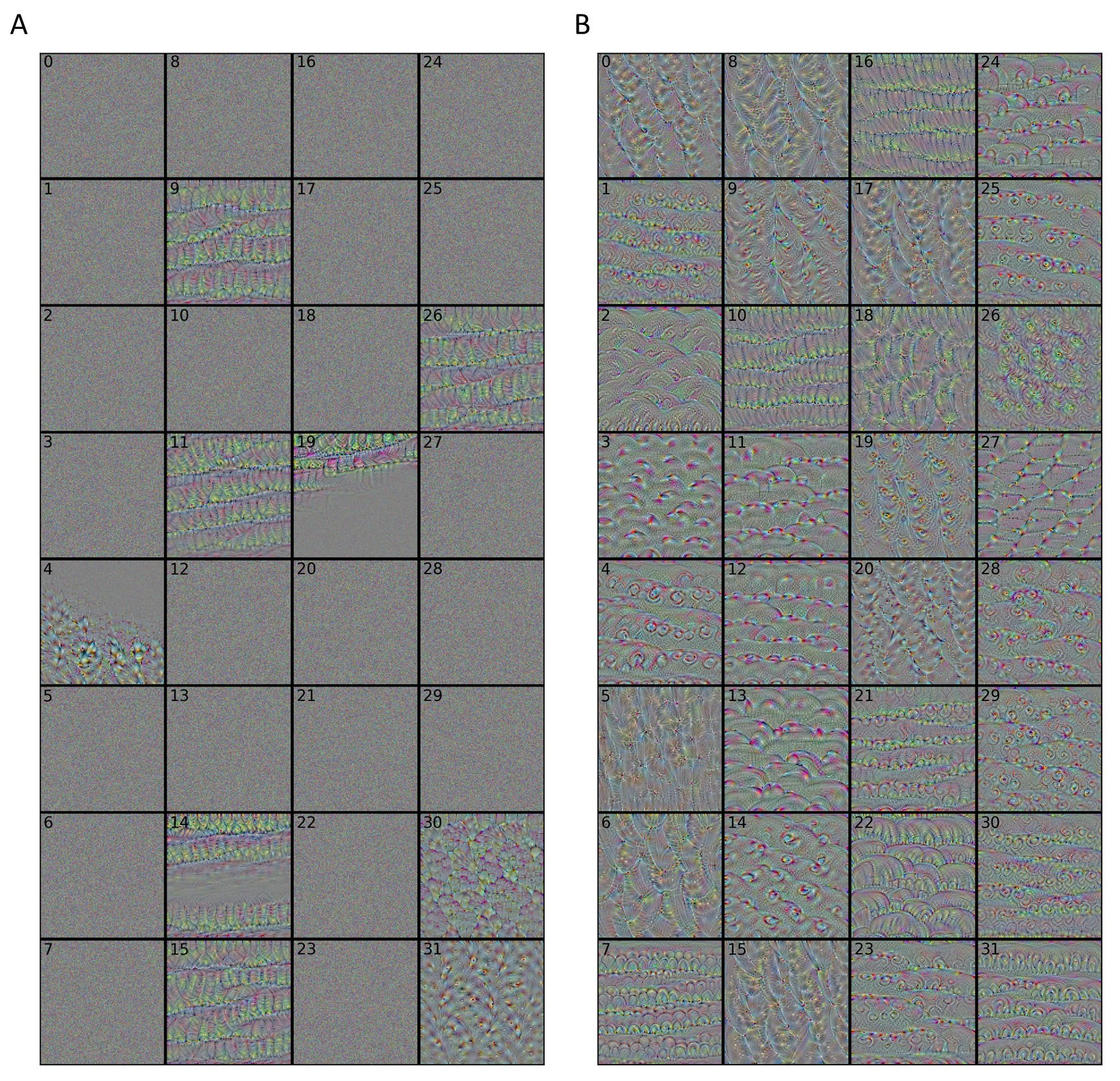}
\caption{Filter patterns of last convolutional layer, generated using gradient ascent method. VGG16 was retuned using \textit{small images} of keratocysts and non-keratocysts. Numbers denote filter numbers. A) Kernel parameters obtained using ImageNet were employed as initial values of the CNN. B) Trainable parameters were initialized using Gaussian distribution before refitting.}
\label{fig5:view}
\end{figure}

Using a \textit{small image}-sized window, \textit{large images} were scanned, and feature maps representing the probability of \textbf{K} in each grid were produced (Figure 6). The scan strides were determined such that approximately one-third of the adjacent windows overlapped, yielding 20$\times$16 grids of the cropped patches. The patches were rotated by 0 to 360\textdegree\ degrees in steps of 15\textdegree, and the margins were filled with white. Further, all the rotated patches were analyzed using the trained first CNN in order to calculate the probability (Figure 6A). A sample probability map with three dimensions (angle$\times$height$\times$width) is shown as a heatmap in Figure 6B.  The probability in each grid was maximal when the rotation angle matched the orientation of the training images where the baso-apical axis was vertical and the cystic space was located on the upper side. In addition, the probability was relatively high for a 180\textdegree\ inversed rotation and very low for all other angles; this was indicative of the orientation dependency of the model. The sensitivity of \textbf{K} detection was satisfactory. However, the first CNN, which was trained using only \textbf{K} and \textbf{N} images, often returned a high probability for \textbf{K} in response to stromal tissue, particularly in the case of those accompanied by inflammation (Figure 6C). To eliminate this noise, we prepared a \textit{small image} dataset consisting only of images of stromal tissue (see Figure 2D) and trained the CNN again, with the aim of categorizing \textit{small images} into 3 classes: \textbf{K}, \textbf{N}, and \textbf{S}. The AUC value for binary classification into \textbf{K} or \textbf{N}, with \textbf{S} being classified as \textbf{N}, using the trained model was sufficiently high (Table 1), and the noise in the feature map for \textit{large images} was reduced by using the model trained for three-category (\textbf{K}, \textbf{N}, or \textbf{S}) classification as compared to that for the model trained for binary \textbf{K} or \textbf{N} classification (Figure 6C). When \textit{small images} of \textbf{N} and \textbf{S} were combined to train the model for classifying the samples into two classes, \textbf{K} or (\textbf{N} and \textbf{S}), the AUC for the \textit{small image} classification of \textbf{K} and \textbf{N} was similar to that for the model trained for \textbf{K}, \textbf{N}, and \textbf{S} (Table 1). However, the feature map of \textit{large images} contained more noise. Because the AUC values for \textit{small image} classification using VGG16 and Resnet34 were similar, we chose to use the VGG16 model trained for three-category (\textbf{K}, \textbf{N}, or \textbf{S}) classification as the first CNN for the rest of the study. To reduce the computational load, a step size of 30\textdegree\ was used for the rotation angles.

\begin{figure}[!hbt]\centering
\captionsetup{format = plain, justification=justified}
\includegraphics[width=9cm, bb=0 -20 1645 1152]{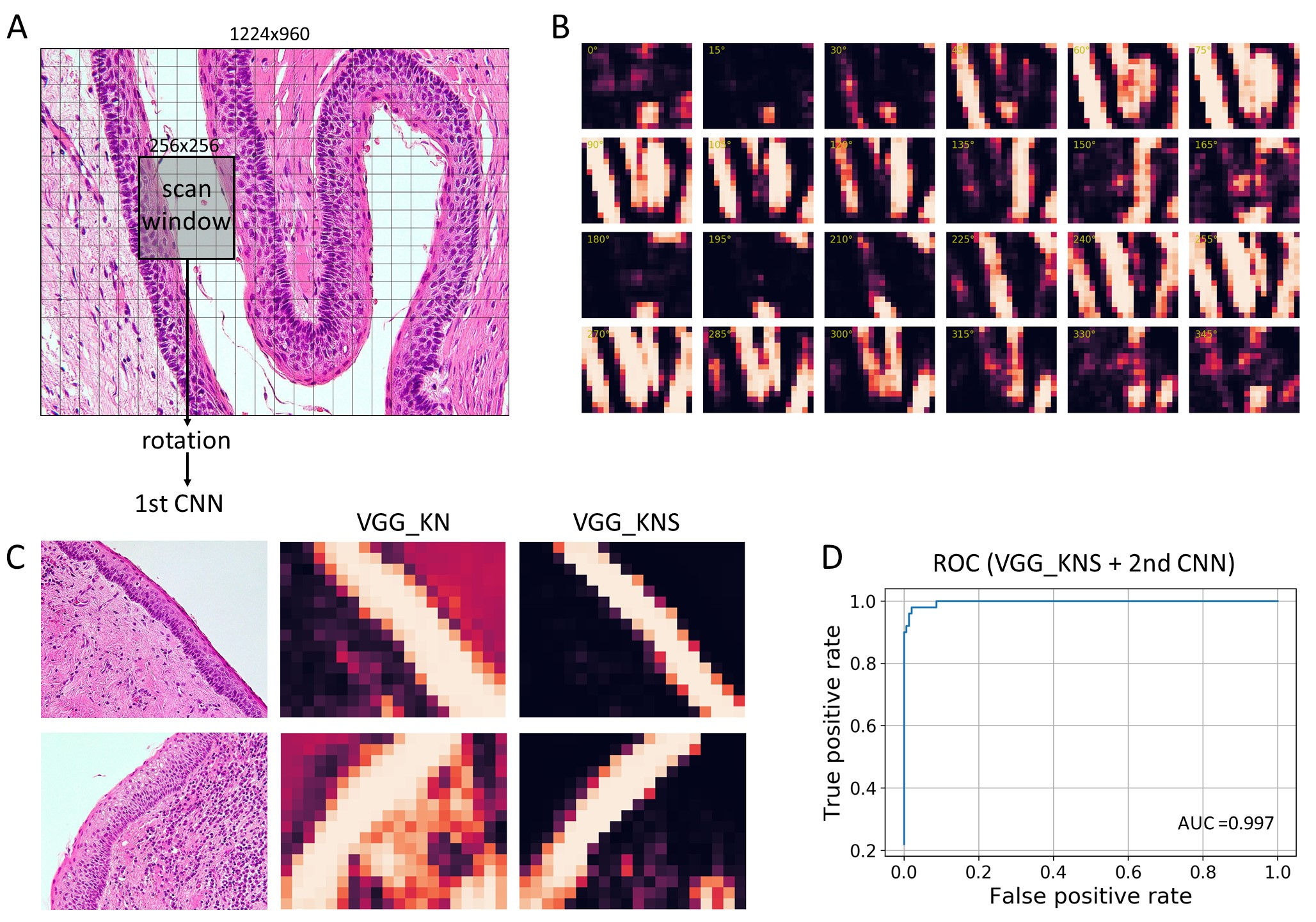}
\caption{Analysis of \textit{large images} using patch classification. A) Figure shows scales for \textit{large images}, \textit{small image}-sized scan window, and scan strides, which are depicted by cross-cut lines. B) Feature map of A) was generated using trained first CNN (VGG16) and is displayed as heatmap of probability for keratocysts. Each tile corresponds to different rotation angle of patches. C) Heatmap of maximum probability at each spatial tile, providing rough picture of keratocyst architecture. VGG\_KN was trained using dataset with binary (keratocysts (\textbf{K}) and non-keratocyst (\textbf{N})) classification. VGG\_KNS was trained using dataset with three-category (\textbf{K}, \textbf{N}, and stroma (\textbf{S})) classification. D) ROC curve and AUC of model (VGG\_KNS + 2nd CNN) obtained using dataset composed of 50 \textit{large images} each of \textbf{K}, \textbf{N}, \textbf{S}, and radicular cysts (\textbf{R}).}
\label{fig6:view}
\end{figure}

A total of 658 \textit{large images} of \textbf{K} and 608 \textit{large images} of \textbf{N} were used to train the second CNN. These were scanned and analyzed using the first CNN in order to generate feature maps with four dimensions (12$\times$16$\times$20$\times$3, (angle$\times$height$\times$\\width$\times$class)), which were then fed into the second CNN for training with the aim of classifying the entire \textit{large image} dataset into \textbf{K} or \textbf{N}. A simple network with three convolutional layers was used for the second CNN.

Finally, the entire model (the trained first and second CNNs) was evaluated using the test dataset, which consisted of 50 \textit{large images} each of \textbf{K}, \textbf{N}, \textbf{S}, and \textbf{R} (Figure 7). While images of \textbf{R} were not included in the training data, we expected that the areas with inflammation would be classified as \textbf{S} and that the areas without inflammation would be classified as \textbf{N}, given the histological similarity between them. In both cases, it was expected that the images would be classified as belonging to the non-keratocyst category. The test dataset was prepared using the cases that had not been included in the training dataset. During the identification of \textbf{K} in 200 \textit{large images}, the AUC was 0.997 (Figure 6D). For a threshold of 0.5, the accuracy, recall (sensitivity), precision, and specificity were 0.980, 0.980, 0.942, and 0.980, respectively.

\begin{figure*}[!hbt]%\centering
\captionsetup{format = plain, justification=justified}
\includegraphics[width=8.5cm, bb=0 -20 1645 1252]{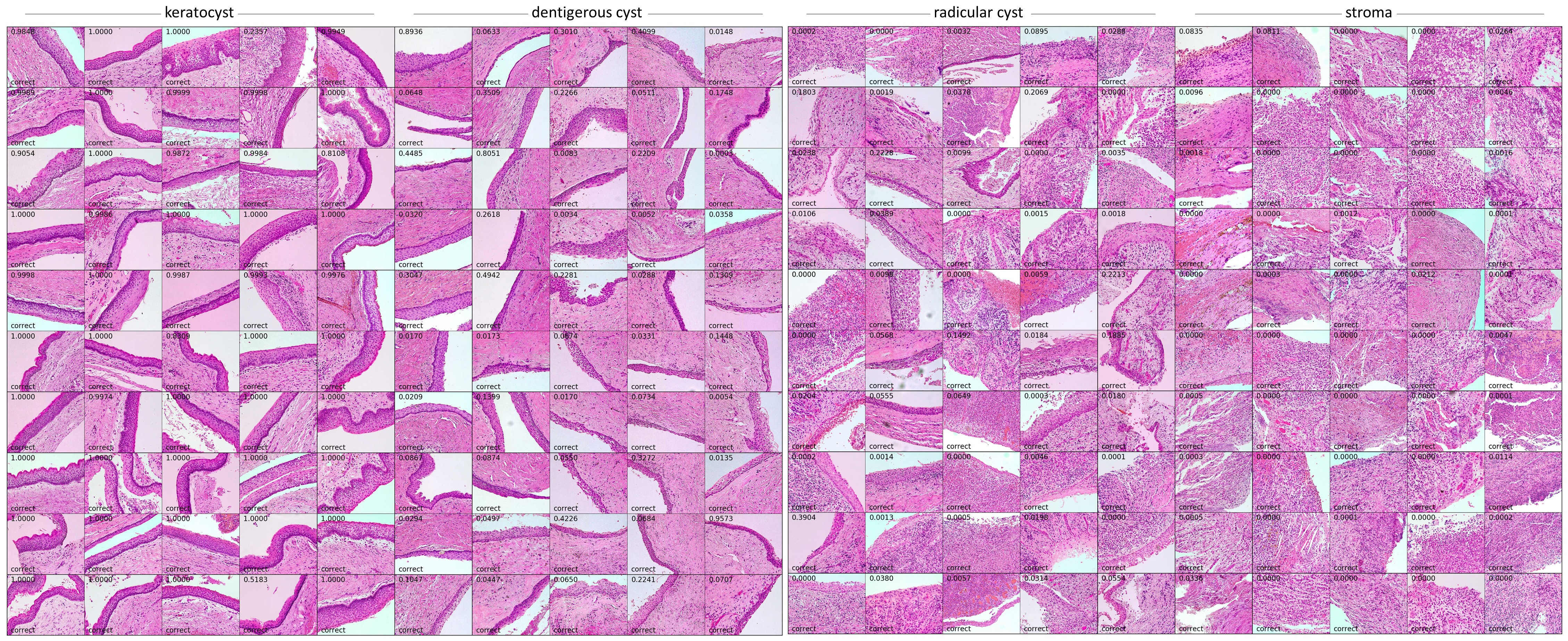}
\caption{Test dataset and predictions made by model. Numbers represent probabilities of keratocysts. Correct predictions are labelled (threshold = 0.5).}
\label{fig7:view}
\end{figure*}

\section*{Discussion}
The \textit{small image}-sized patch can be considered the smallest unit based on which the keratocyst-specific histology can be defined. If it were to be divided into smaller patches, the resulting patches would no longer retain the attributes characteristic of keratocysts. Thus, this patch classification strategy of using two CNNs, as was done in this study, is a reasonable one for identifying keratocysts. The process involved overcoming the three challenges described in the Introduction section. The first CNN investigated the cellular features and their arrangement, while the second CNN investigated the cyst architecture and assigned a class to the entire image. The trained model could successfully recognize keratocysts in most instances. However, several challenges remain with respect to the clinical applicability of the proposed method. False negatives arose mainly owing to the omission of training data corresponding to cases with a very thick lining epithelium. This was a compromise made to simplify the classification task, and most false negatives can be avoided simply by increasing the patch size to encompass the thickest lining epithelium. In 99\% of the cases used, the lining epithelium was less than 600 px thick, with the thickest part being approximately 800 px in thickness. Increasing the size of the images in the dataset would mean a tradeoff with the number of images. Further, a bigger window would contain more stromal areas and may decrease the training efficiency. The use of both small and large windows together (for example, 256$\times$256 and 600$\times$600 px windows) could be one workaround to avoid this tradeoff. False positives were reported in the case of distorted lining epithelia owing to the artifacts arising during tissue processing. Since we probably avoided these ``dirty" parts during patch collection by instinct, adding these ``bad" images to the training dataset may improve the specificity.

A completely automated system that can examine images of the entire slide taken using a virtual slide system needs to be developed. For this, \textit{large images} would have to be analyzed. For example, a preliminary survey showed that 10 randomly chosen biopsy specimens of keratocysts had tissue areas corresponding to an average of 248 (range of 182--411) \textit{large images}. The algorithms used in this study could analyze one \textit{large image} in approximately 20 s on a single-GPU computer. Thus, the total computation time for a single specimen would be approximately 1--2 h, if the tissue areas were to be extracted during preprocessing and divided into \textit{large image}-sized patches for stepwise analysis. This process would take significantly longer than what pathologists take to analyze a single specimen (a few minutes or less). However, this should not be an obstacle to the practical implementation of the proposed system, since pathological investigations do not require processing times of the order of seconds or minutes. Further, the computation time could be decreased by optimizing the algorithm and using parallel processing. The lining epithelium is usually present only in a limited area. For example, in the abovementioned specimens, the lining epithelia were included only in approximately 10--20\% of \textit{large images}. Theoretically, only one area with a definitive keratocyst feature is sufficient for diagnosis. However, the presence of only a few positives in the \textit{small image} patches will lead to a high number of false positives unless the specificity of the first CNN is improved further. Hence, having one or more positives in the \textit{large image} tiles would be a practical solution. The accuracy of the ``whole slide" image analysis process could probably be improved by adding a third CNN that evaluates the feature map composed of the outputs from the second CNN. However, the number of data samples cannot exceed the number of specimens used for training the third CNN; this limits the size of the dataset. ``Whole slide" image analysis requires the optimization and refinement of the model as well as the collection of more data as the next step.

The process of decision making using neural networks is commonly regarded as one involving a black box. Even if the correct answer is obtained, a judgment based on features not related to the pathophysiology would not be acceptable to clinicians because of the lack of accountability. The activation pattern of the intermediate layer suggested that the first CNN was responding to cellular features similar to those that pathologists pay close attention to when examining keratocysts. The ``capsule pattern" filters seemed to respond to the aligned elliptical nuclei of the basal cells, which are a definitive feature of keratocysts, suggesting that the CNN spontaneously learned of this fundamental diagnostic criterion. However, the promising results obtained can be attributed to the favorable nature of keratocysts, which have a stereotypic appearance. In case of other diseases that usually exhibit different histologies, the extraction of the features essential for classification would be a more demanding task.

Finally, filters that responded strongly to the spinous cells with prominent nucleoli were also generated. The outputs of these filters were significantly higher in the case of keratocysts. In a biological context, this seemingly reflects the high protein synthesis activity in the spinous layer of keratocysts, given the high degree of keratinization. Although this fact may be self-evident, it is not mentioned in textbooks and is an example of a neural network discovering a histological feature that was overlooked by human observers.

Recognizing keratocysts in \textit{small images} was mainly dependent on a few types of filters in the last convolutional layer. This simplified optimization resulted in a high overall accuracy but led to failure in distinguishing samples with poor histological characteristics. The palisading pattern is almost absent in some keratocysts, whereas non-keratocysts may exhibit a pseudo-palisading pattern. Pathologists can also discern other features such as the degree of keratinization and cell attachment. Although these are not crucial for the classification decision, pathologists use them as supporting evidence. However, most of these features seem to have been ignored by the CNNs. Ideally, an automated diagnosis system should be trained to pay attention to the same histologic features as pathologists because they may contain information relevant to the pathogenesis of the lesions. A straightforward solution for ensuring that the neural network learns these specific histological characteristics would be to annotate the images with the particular features as subclasses. Most of these attributes are based on subtle differences in the texture or architecture and are not easy to label. Additional studies are needed to determine whether CNNs can handle these ambiguous histological features.

In practice, additional information apart from that obtained by H\&E staining is available for pathological diagnosis. Histopathology was initially based on H\&E staining. However, there are other methods now available that are more suitable for machine vision. For example, the palisading patterns can be recognized using a simple nuclear staining technique such as DAPI staining. This can also reduce the size of the dataset because the DAPI-stained images consist of one-dimensional density datapoints. In addition, the cyst architecture can be visualized by immunostaining using an anti-keratin antibody instead of based on epithelial segmentation. In the case of keratocysts and non-keratocysts classification, immunostaining using the anti-BCL2 antibody would be useful, since it specifically labels the basal cells of keratocysts~\cite{bib26}. With the use of several alternative diagnostic markers, automated pathological diagnosis may become more feasible based on the use of the appropriate immunostaining method instead of H\&E staining, if one can take care of the higher costs involved. That said, it remains a challenge to develop an image analysis system that can mimic the performance of human experts in analyzing H\&E-stained tissue specimens.

%------------------------------------------------
\phantomsection
\section*{Acknowledgments} % The \section*{} command stops section numbering

\addcontentsline{toc}{section}{Acknowledgments} % Adds this section to the table of contents

This work was supported by JSPS KAKENHI [grant number JP16K11438]. The funders had no role in study design, data collection and analysis, decision to publish, or preparation of the manuscript.


\begin{thebibliography}{10}

\bibitem{bib1}
Litjens G, Kooi T, Bejnordi BE, Setio AAA, Ciompi F, Ghafoorian M, van der Laak JAWM, van Ginneken B, Sánchez CI.
\newblock {{A} survey on deep learning in medical image analysis}
\newblock Med Image Anal. 2017;42: 60-88.

\bibitem{bib2}
Shen D, Wu G, Suk HI.
\newblock {{D}eep Learning in Medical Image Analysis.}
\newblock Annu Rev Biomed Eng. 2017;19: 221-248.

\bibitem{bib3}
Fuyong X, Yuanpu X, Hai S, Fujun L, Lin Y.
\newblock {{D}eep Learning in Microscopy Image Analysis: A Survey.}
\newblock IEEE Trans Neural Netw Learn Syst. 2018;29: 4550-4568.

\bibitem{bib4}
Ker J, Wang L, Rao J, Lim T.
\newblock {Deep Learning Applications in Medical Image Analysis.}
\newblock IEEE Access. 2017;6: 9375-9389.

\bibitem{bib5}
Müller AC, Guido S.
\newblock {Introduction to Machine Learning with Python.}
\newblock 1st ed. 2017. O'Reilly Japan.

\bibitem{bib6}
Couture HD, Williams LA, Geradts J, Nyante SJ, Butler EN, Marron JS, Perou CM, Troester MA, Niethammer M.
\newblock {Image analysis with deep learning to predict breast cancer grade, ER status, histologic subtype, and intrinsic subtype.}
\newblock NPJ Breast Cancer. 2018;4: 30-018-0079-1.

\bibitem{bib7}
Feng Y, Zhang L, Yi Z.
\newblock {Breast cancer cell nuclei classification in histopathology images using deep neural networks.}
\newblock Int J Comput Assist Radiol Surg. 2018;13: 179-191.

\bibitem{bib8}
Maqlin P, Thamburaj R, Mammen J, Manipadam M.
\newblock {Automated nuclear pleomorphism scoring in breast cancer histopathology images using deep neural networks.}
\newblock Mining Intelligence and Knowledge Exploration. (Lecture Notes in Computer Science). 2015;9468: 269-76. Springer International Publishing.

\bibitem{bib9}
Arvaniti E, Fricker KS, Moret M, Rupp N, Hermanns T, Fankhauser C, Wey N, Wild PJ, Rüschoff JH, Claassen M.
\newblock {Automated Gleason grading of prostate cancer tissue microarrays via deep learning.}
\newblock Sci Rep. 2018;8: 12054-018-30535-1.

\bibitem{bib10}
Diamond J, Anderson NH, Bartels PH, Montironi R, Hamilton PW.
\newblock {The use of morphological characteristics and texture analysis in the identification of tissue composition in prostatic neoplasia.}
\newblock Hum Pathol. 2004;35: 1121-1131.

\bibitem{bib11}
Bychkov D, Linder N, Turkki R, Nordling S, Kovanen PE, Verrill C, Walliander M, Lundin M, Haglund C, Lundin J.
\newblock {Deep learning based tissue analysis predicts outcome in colorectal cancer.}
\newblock Sci Rep. 2018;8: 3395-018-21758-3.

\bibitem{bib12}
Sirinukunwattana K, Raza SEA, Tsang Y, Snead DRJ, Ian A.
\newblock {Locality Sensitive Deep Learning for Detection and Classification of Nuclei in Routine Colon Cancer Histology Images.}
\newblock IEEE Transactions on Medical Imaging. 2016;35: 1196-1206.

\bibitem{bib13}
Jones AV, Craig GT, Franklin CD.
\newblock {Range and demographics of odontogenic cysts diagnosed in a UK population over a 30-year period.}
\newblock J Oral Pathol Med. 2006;35: 500-507.

\bibitem{bib14}
Shear M.
\newblock {Developmental odontogenic cysts. An update.}
\newblock J Oral Pathol Med. 1994;23: 1-11.

\bibitem{bib15}
Shear M.
\newblock {The aggressive nature of the odontogenic keratocyst: is it a benign cystic neoplasm? Part 1. Clinical and early experimental evidence of aggressive behaviour.}
\newblock Oral Oncol. 2002;38: 219-226.

\bibitem{bib16}
Tolstunov L, Treasure T.
\newblock {Surgical treatment algorithm for odontogenic keratocyst: combined treatment of odontogenic keratocyst and mandibular defect with marsupialization, enucleation, iliac crest bone graft, and dental implants.}
\newblock J Oral Maxillofac Surg. 2008;66: 1025-1036.

\bibitem{bib17}
Landini G.
\newblock {Quantitative analysis of the epithelial lining architecture in radicular cysts and odontogenic keratocysts.}
\newblock Head Face Med. 2006;2: 4-160X-2-4.

\bibitem{bib18}
Han JW, Breckon T, Randell D, Landini G.
\newblock {Radicular cysts and odontogenic keratocysts epithelia classification using cascaded Haar classifiers.}
\newblock  2008.

\bibitem{bib19}
Florindo JB, Bruno OM, Landini G.
\newblock {Morphological classification of odontogenic keratocysts using Bouligand-Minkowski fractal descriptors.}
\newblock Comput Biol Med. 2017;81: 1-10.

\bibitem{bib20}
Chollet F.
\newblock {Keras, GitHub. https://github.com/fchollet/\\keras.} 
\newblock 

\bibitem{bib21}
Simonyan K, Zisserman A.
\newblock {Very Deep Convolutional Networks for Large-Scale Image Recognition.}
\newblock arXiv:1409.1556 [cs.CV].

\bibitem{bib22}
He K, Zhang X, Ren S, Sun J.
\newblock {Deep Residual Learning for Image Recognition.}
\newblock arXiv:1512.03385 [cs.CV].

\bibitem{bib23}
Russakovsky O, DengHao J, Krause J, Satheesh S, Ma S, Huang Z, et al.
\newblock {ImageNet Large Scale Visual Recognition Challenge.}
\newblock International Journal of Computer Vision. 2015;115: 211-252.

\bibitem{bib24}
Chollet F.
\newblock {Deep Learning with Python.}
\newblock 1st ed. 2017. Manning Publications.

\bibitem{bib25}
Selvaraju RR, Cogswell M, Das A, Vedantam R, Parikh D, Batra D.
\newblock {Grad-CAM: Visual Explanations from Deep Networks via Gradient-based Localization.}
\newblock arXiv:1610.02391v3 [cs.CV].

\bibitem{bib26}
Piattelli A, Fioroni M, Rubini C.
\newblock {Differentiation of odontogenic keratocysts from other odontogenic cysts by the expression of bcl-2 immunoreactivity.}
\newblock Oral Oncol. 1998;34: 404-407.
\end{thebibliography}
\end{document}